\documentclass[11pt]{article}
\usepackage[T1]{fontenc}
\usepackage[preprint]{acl}
\usepackage{times}
\usepackage{latexsym}

\usepackage{amsmath,amsfonts,bm}

\def\eqref#1{equation~\ref{#1}}

\def\1{\bm{1}}

\DeclareMathAlphabet{\mathsfit}{\encodingdefault}{\sfdefault}{m}{sl}
\SetMathAlphabet{\mathsfit}{bold}{\encodingdefault}{\sfdefault}{bx}{n}

\usepackage{graphicx}
\usepackage{booktabs}
\usepackage{array}
\usepackage{longtable}
\usepackage{subcaption}
\usepackage{float}
\usepackage{xcolor}
\usepackage{colortbl}
\usepackage{placeins}
\usepackage{hyperref}
\usepackage{url}
\usepackage{multirow}   
\usepackage{enumitem}
\makeatletter
\renewcommand{\outauthor}{%
  \begingroup
  \def\And{\unskip\hspace{1.4em}\ignorespaces}%
  \def\AND{\unskip\\[0.35em]\bfseries\ignorespaces}%
  \begin{tabular}[t]{@{}c@{}}%
    \ifacl@anonymize
      \bfseries Anonymous ACL submission%
    \else
      \bfseries\@author
    \fi
  \end{tabular}%
  \endgroup
}
\makeatother

\title{BuySideFlow: A Point-in-Time Benchmark for LLM Agents in Investment Research}

\author{
Dong Bo \And
Fanjin Zhang \And
Ke Cheng \And
Xinxin Zhao \And
Xiaowei Su
\AND
Yang Wang \And
Jinghang Xu \And
Cuiping Li \And
Hong Chen \And
Jing Zhang
}

\newcommand{\hide}[1]{}

\begin{document}
\raggedbottom

\maketitle

\begin{abstract}

Large language models (LLMs) can generate SQL and SQL--Python workflows for investment research, but ensuring point-in-time (PIT) validity remains challenging. Agents must enforce disclosure cutoffs and select the appropriate historical states and record versions throughout data retrieval and financial analysis. We introduce \textsc{BuySideFlow}, a benchmark of 408 practitioner-derived Chinese-language tasks spanning equities, funds, fixed income, and macroeconomics over a fixed snapshot from a licensed production financial database. We provide executable references and a taxonomy of temporal errors, pairing result checks with expert audits of executed workflows. We evaluate nine agent configurations on all tasks, with result accuracy ranging from 16.2\% to 36.8\%. An audit of 900 runs on 100 shared tasks confirms PIT violations in 193 of the 380 runs that pass result checks (50.8\%), showing that output agreement alone does not establish temporal validity.

\hide{
Investment research requires SQL and SQL--Python workflows to respect what was available and which historical states and data versions applied at each decision time. We introduce \textsc{BuySideFlow}, a point-in-time (PIT) workflow benchmark for Chinese buy-side investment research. Its 408 practitioner-derived tasks span equities, funds, fixed income, and macroeconomic analysis over a sample snapshot drawn from a licensed production financial database. The benchmark translates disclosure, historical-state, and version-selection requirements into task-specific criteria for evaluating LLM-generated workflows. Execution-result checks measure task completion, while a human audit examines executed dependencies, accepts equivalent implementations, and retains judgments with insufficient evidence as unresolved. We evaluate nine agent configurations on all 408 tasks and audit 900 configuration--task records from a shared subset of 100 tasks. The expert audit identifies at least one PIT violation in 193 of the 380 records that pass result checks (50.8\%). These errors concern when information became available, which historical state applied, and how record versions were selected.
}
\end{abstract}

\section{Introduction}
\label{sec:introduction}

Text-to-SQL enables natural-language access to databases, and large
language model (LLM) agents extend this capability to analytical workflows
\citep{lei2025spider2}. In financial and quantitative research, such
workflows can help analysts assess companies and evaluate investment
strategies. Because these analyses inform investment decisions, they
require careful data selection and accurate calculations
\citep{chen2021finqa}. When reconstructing historical investment decisions,
the analysis must also reflect what information was available at each
decision date. Incorporating later disclosures or revisions can distort
the conclusions and overstate a strategy's historical performance
\citep{croushore2001realtime,daniel2008look}. Careful treatment of time
is therefore essential to reliable automation of investment research.

\hide{
Text-to-SQL enables users to query databases in natural language. Spider and BIRD evaluate SQL generation across domains \citep{yu2018spider,li2023can}, while Spider~2.0 evaluates enterprise workflows in which large language model (LLM) agents consult documentation, inspect schemas, and execute SQL and Python \citep{lei2025spider2}. When such workflows assess what could have been concluded at a historical decision time, the information available then is part of the task. Later disclosures or revisions can change the evidence underlying a result.
}

Existing benchmarks cover financial SQL generation, enterprise analytical
workflows, and financial agents that use multiple tools
\citep{zhang2024finsql,lei2025spider2,zeng2025fingaia}.
Recent work also checks dependence on future data and restricts agent
inputs to information available at the relevant time
\citep{luo2026alphaqt,xiao2026financeharness,cai2026openpm}.
Evaluating generated SQL and SQL--Python workflows for historical
investment research presents two connected challenges.
First, investment research combines data retrieval, financial calculations, and task-specific outputs, requiring \textbf{evaluation of complete workflows.}
Second, \textbf{point-in-time (PIT) validity requires respecting information
cutoffs and selecting the historical states and record versions required
by each task.}

\begin{figure*}[!t]
\centering
\includegraphics[width=\linewidth]{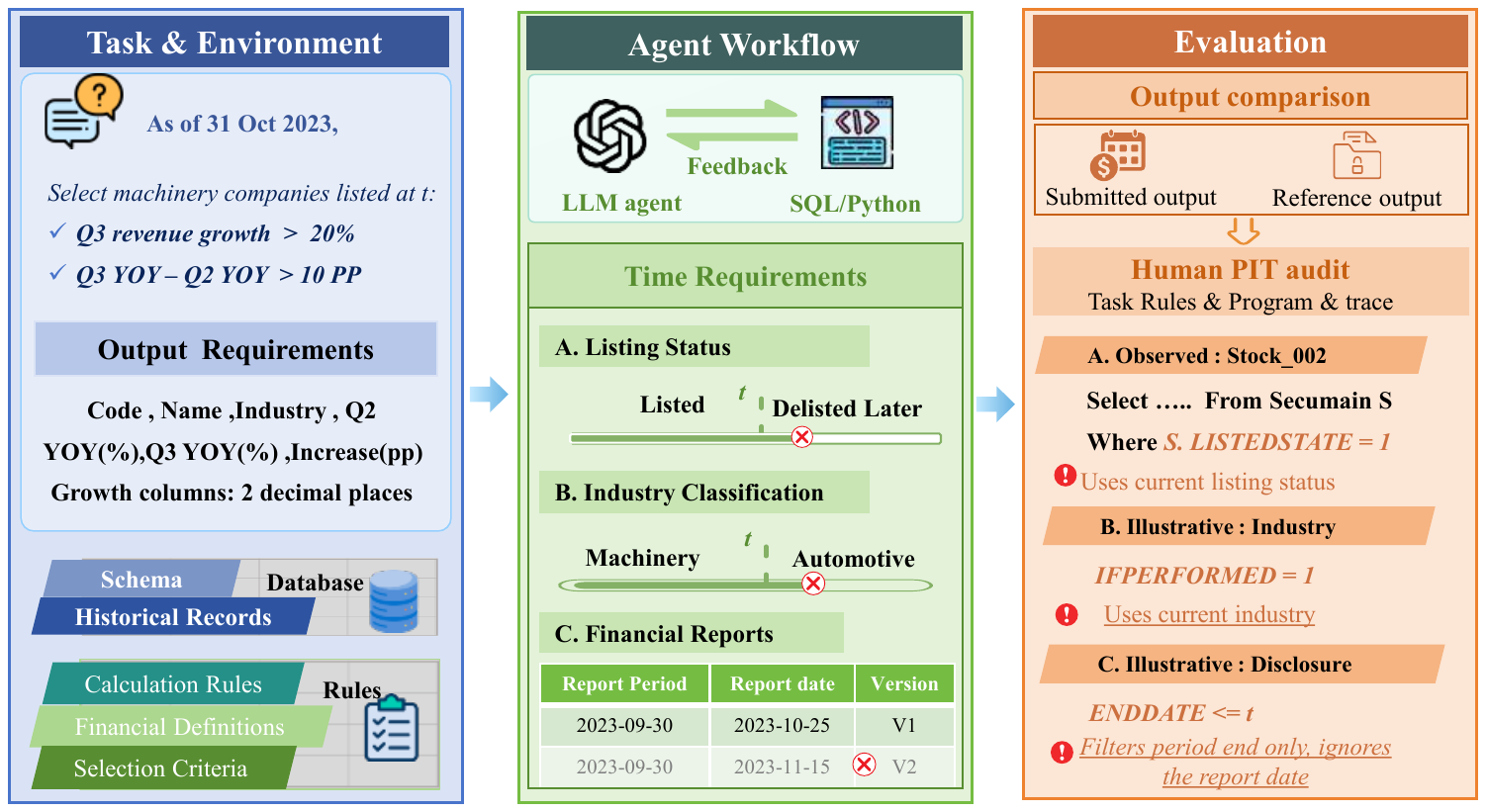}
\caption{Overview of BuySideFlow. Agents generate SQL or SQL--Python workflows for historical investment research. Outputs are compared with reference results. Point-in-time (PIT) audits check what information was available at the research date and whether the correct historical company status, industry classifications, and report versions were used. A shows an observed error; B--C and the timelines are illustrative. YoY denotes year-on-year growth; pp denotes percentage points.}
\label{fig:buysideflow_pipeline}
\end{figure*}

For example, in the selection task of historical companies in Figure~\ref{fig:buysideflow_pipeline}, PIT violations include using a financial report disclosed after the research date, even if it covers an earlier period, or using current listing status or industry classifications when historical values are required.

\hide{
Investment research makes these requirements concrete. A financial report may cover a period ending before the research date yet remain unpublished, while later revisions can change previously released values \citep{croushore2001realtime}. A company's current listing status or industry classification may also differ from the one in effect at that date. We call compliance with the task's information cutoff, historical-state, and record-version requirements \emph{point-in-time (PIT) validity}.

Our empirical question is whether LLM agents generate SQL and SQL--Python workflows that satisfy both the requested outputs and these temporal requirements. Figure~\ref{fig:buysideflow_pipeline} gives an overview using a historical company-selection task. The program must recover listing status and industry classification at the research date and select financial report versions disclosed by then. Result comparison checks the outputs; human PIT auditing examines temporal data choices across SQL and Python.
}

To deal with these challenges, we introduce \textsc{BuySideFlow}, a benchmark of 408 Chinese research questions adapted from buy-side research practice. Agents
generate SQL or SQL--Python workflows over a fixed sample drawn from a
licensed production financial database. We evaluate nine LLM--agent
configurations on all tasks and audit 900 submissions from a shared
subset of 100 questions, assessing both task completion and PIT validity.
Our contributions are:

\begin{itemize}[leftmargin=*]
   \item \textbf{A benchmark grounded in investment research.}
    The tasks span equities, funds, fixed income, and macroeconomics,
    supporting SQL and SQL--Python workflows with tables, figures, and
    mixed outputs. Task instructions, executable references, and outputs
    undergo multiple rounds of expert review, execution checks, and
    revision to assess financial correctness and compliance with temporal
    requirements.
    \item \textbf{PIT taxonomy and workflow evaluation.}
    We provide a taxonomy of PIT violations covering errors in time
    interpretation, substitution of current states for historical ones,
    and faulty historical reconstruction. We pair comparisons against
    reference outputs with expert audits. Each confirmed violation is linked to a task requirement and
    execution evidence.

    \item \textbf{Systematic evaluation and temporal error analysis.}
    Result accuracy across the nine configurations ranges from
    16.2\% to 36.8\%. Among the 380 audited submissions that pass result
    checks, 193 (50.8\%) contain confirmed PIT violations. These findings
    show that completing investment research tasks remains challenging
    and that output agreement alone does not establish temporal validity.
\end{itemize}

\section{\textsc{BuySideFlow}}
\label{sec:benchmark}

\subsection{Tasks and Research Environment}
\label{sec:schema_release}

\paragraph{Task specification.}
Each task provides an investment research question, optional structured
inputs, and output requirements. Together, these specify what to analyze,
which data to include, the relevant dates, and the rules for financial
calculations. Structured inputs may provide dates, analysis periods, or
portfolios. Output requirements specify formatting details such as fields, units,
numerical precision, and ordering. Tasks may involve a single historical
date, a sequence of portfolio adjustment dates, or a period around an event.

\paragraph{Research database.}
All evaluations use the same fixed sample snapshot of the licensed Gildata
financial database. The sample contains 110 tables and 4,547 columns,
including tables for security information, listing histories, industry
and fund classifications, and financial-statement data. Programs select
task-specific historical records from this shared database. Appendix~\ref{app:construction_audit_details}
describes access, retained data, and task-level validation evidence.
Figure~\ref{fig:evaluation-protocol}a summarizes how
agents interact with the database.

\paragraph{Research workflows.}
The tasks cover equities, funds, fixed income, and macroeconomics.
For example, one task groups stocks by profitability and
compares their subsequent returns, while another compares
fund returns and declines from previous peaks. Other tasks rank bonds
by the difference between their yields and government-bond yields,
or identify periods of at least three months when
consumer price indicators remain above producer price indicators. The reference workflows use SQL or SQL with Python.
In \emph{SQL-native} tasks, SQL performs filtering, joins,
aggregation, and financial calculations; Python may wrap execution or save results.
In \emph{SQL--Python} tasks, Python also
coordinates multiple query steps, applies task-specific financial rules,
performs statistical analysis, or generates plots and summary tables.
These labels describe the reference workflows and do not restrict
equivalent agent implementations.

\subsection{Point-in-Time Requirements}
\label{sec:pit_requirements}

\paragraph{Dates in financial data.}
Financial records contain several dates with different meanings.
The \emph{reporting-period end date} identifies the end of the period
covered by a financial statement (e.g., \texttt{ENDDATE}).
The \emph{disclosure date} indicates when the information was made
public (e.g., \texttt{INFOPUBLDATE}).
\emph{Effective dates and validity intervals} describe when a state,
such as an industry classification or a fund manager's appointment,
applies. \emph{Database timestamps} record when a row is inserted or
updated. These dates are not interchangeable
\citep{croushore2001realtime,jensen1999temporal}: \textbf{a period end or
effective date does not establish public disclosure}. For example, a fund manager's appointment takes effect on 23 December but is disclosed on 25 December. A query based on information disclosed by 24 December cannot rely on this record, even though the appointment is already in effect. Insertion time
applies only when the task requires the record to have been in the
database by the cutoff or explicitly permits it as a proxy for
availability. A last-update timestamp does not establish when the
information first became available.
Appendix~\ref{app:temporal_fields} provides the corresponding sources
and field definitions.

\paragraph{Point-in-time constraints.}
Each task specifies the historical dates to analyze and the information cutoffs governing the analysis. These need not coincide: a retrospective
summary may use later revisions only when the task explicitly permits
them. For a fixed task version and database snapshot, a workflow is
\emph{PIT-valid} when the following requirements hold.
\textit{First}, every input affecting a historical decision must be available by the relevant cutoff, including inputs used to filter records or choose analytical steps. 
Second, listing status, index membership, and classifications must match the historical state required by the task. 
Third, when multiple versions exist, the workflow must select the task-required version from eligible records. Using only pre-cutoff records is insufficient if the workflow retains an outdated version or drops a record needed to reconstruct the historical state.
These requirements apply separately at every decision date. Later returns may be used to evaluate an earlier selection, but must not influence that selection, grouping, or input features.
Table~\ref{tab:temporal-task-requirements} summarizes these requirements and the evidence used to check them.

\paragraph{Example: industry weights at successive dates.}
Task \texttt{stock\_069} asks which three Shenwan level-1 industries account for the largest total weight in the CSI~300 index at each dated weight snapshot in 2025, through 31 December.
For each snapshot, the workflow retrieves that date's constituents and weights, assigns industries using classifications available and applicable at that date, and sums weights by industry before ranking them. 
For illustration, suppose a company's move from industry A to industry B is disclosed and takes effect in July, 2025. Its March weight must still count toward industry A. 
Reusing December classifications for all snapshots would instead assign that March weight to industry B, incorrectly applying the later classification to an earlier date.
The 31 December endpoint limits which snapshots are requested; it does not permit using year-end classifications for earlier snapshots.
Appendix~\ref{app:stock069_workflow} (Table~\ref{tab:stock069-workflow}) details the corresponding operations and submission evidence.

\hide{
Tasks specify the historical dates to reconstruct and the information cutoffs governing those reconstructions. These need not coincide: later revisions may be used in a retrospective summary only when the task permits them.

For a fixed task version $T$ and database instance $D$, a submitted SQL or SQL--Python workflow $P$ is \emph{PIT-valid} when, at every decision time $t$ required by $T$, its decision-relevant dependencies satisfy the task's information-availability constraints, and its historical-state and version selections follow the task's temporal rules. These dependencies include selection and control flow: each contributing fact $x$ must satisfy $a_T(x)\leq t$, where $a_T$ denotes task-defined availability. Historical selection must also preserve eligible records: using only past information is insufficient if the workflow selects an obsolete version or incorrectly excludes a historical state. Table~\ref{tab:temporal-task-requirements} maps recurring requirements to research operations and submission evidence.
}

\begin{table*}[!htbp]
\centering
\small
\setlength{\tabcolsep}{2.5pt}
\renewcommand{\arraystretch}{1.12}
\caption{Temporal requirements annotated from the released task statements and applicable rules (Appendix~\ref{app:construction_audit_details}). Labels overlap; counts and percentages use all 408 tasks.}
\label{tab:temporal-task-requirements}
\begin{tabular}{@{}l>{\raggedright\arraybackslash}p{0.728\linewidth}r@{}}
\toprule
Requirement & Audit focus & Tasks, $n$ (\%) \\
\midrule
Disclosure cutoff & Task-defined availability dates for selection or calculation inputs, checked against the cutoff & 311 (76.2\%) \\
\addlinespace[3pt]
Historical state & Listing, membership, and classification histories defining the universe at the research date & 272 (66.7\%) \\
\addlinespace[3pt]
Version selection & Version keys and operation order (filtering, ranking, deduplication) for selecting applicable records & 233 (57.1\%) \\
\addlinespace[3pt]
Decision sequence & State reconstruction or grouping at successive dates, checking inputs against each date's information cutoff & 71 (17.4\%) \\
\addlinespace[3pt]
Ex-post labels & Later outcomes for evaluation, without influencing earlier selection, grouping, or features & 58 (14.2\%) \\
\bottomrule
\end{tabular}
\end{table*}

\subsection{Dataset Construction and Quality Control}
\label{sec:construction}
\label{sec:benchmark_statistics}

Two researchers in a securities firm's asset allocation department collected investment research questions.
An asset allocation expert screened them before three portfolio managers reviewed them, one each for equities, bonds, and fund investment.
Their feedback led us to remove similar screening and ranking tasks, add event studies, periodic factor tests and portfolio rebalancing, and clarify candidate universes, calculation conventions and output requirements.
Financial data engineers and quantitative researchers implemented reference workflows and saved their outputs.

\begin{figure*}[t]
\centering
\includegraphics[width=\textwidth]{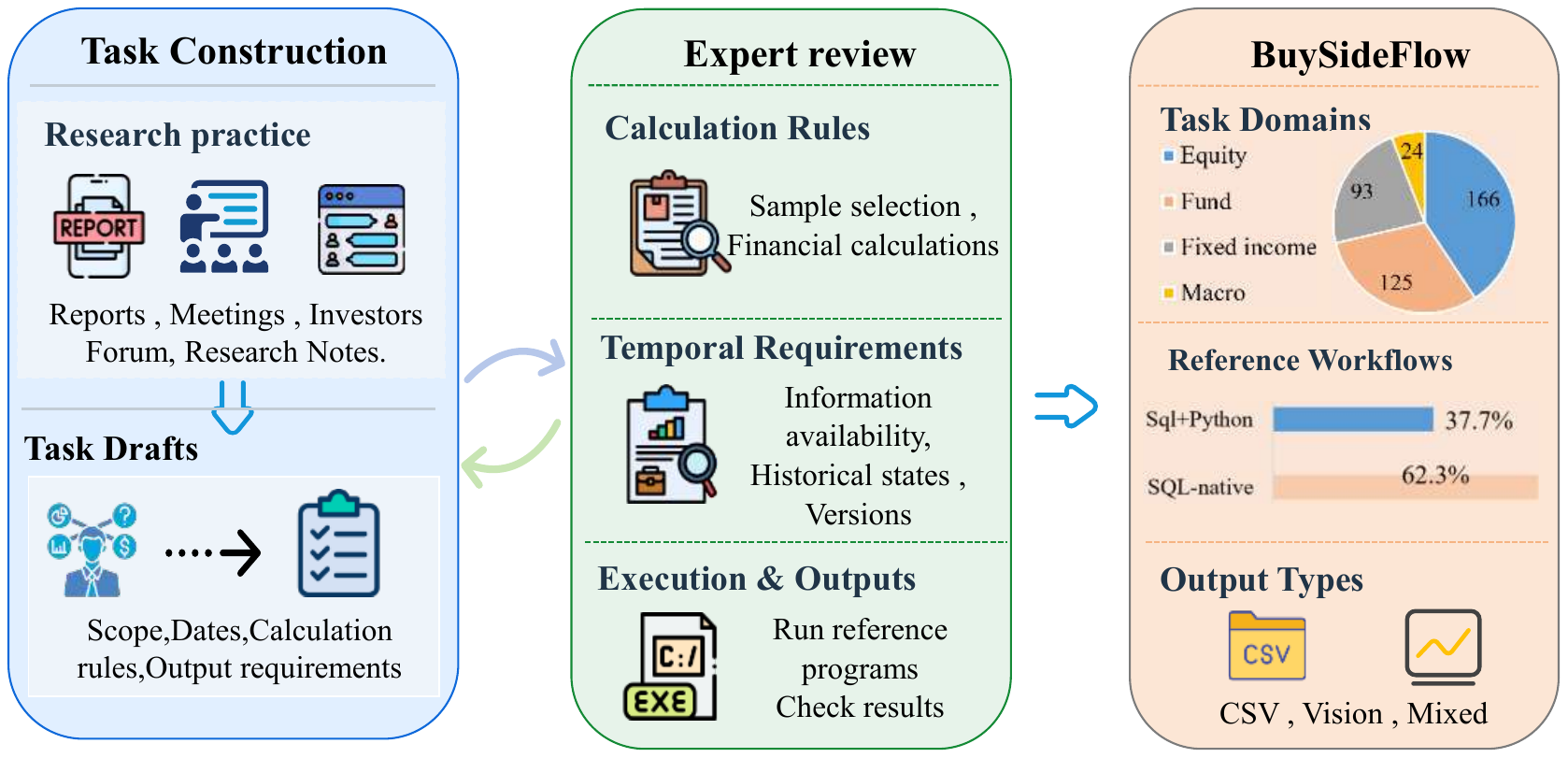}
\caption{Benchmark construction and dataset composition of BuySideFlow. Percentages use all 408 tasks.}
\label{fig:construction-pipeline}
\end{figure*}

A buy-side quantitative researcher and a financial data scientist independently reviewed task statements, reference programs, and outputs. Retained records document four review rounds, the last with no recorded issues (Figure~\ref{fig:construction-pipeline}).
Checks covered financial calculations, disclosure cutoffs, historical states, record versions, and agreement between executed results and output requirements.
The first round covered all tasks; later rounds revisited unresolved issues and checked revisions.
Across rounds, 208 released tasks (51.0\%) had at least one recorded issue.
This construction review preceded the audit of model submissions in Section~\ref{sec:protocol}.

Finally, the current version contains 408 Chinese-language tasks (Figure~\ref{fig:construction-pipeline}): 166 in equities (40.7\%), 125 in funds (30.6\%), 93 in fixed income (22.8\%), and 24 in macroeconomics (5.9\%).
Reference workflows comprise 254 SQL-native tasks (62.3\%) and 154 SQL--Python tasks (37.7\%).
Outputs include 401 CSV tasks, six image-oriented tasks, and one mixed task.
\section{Evaluation Protocol}
\label{sec:protocol}

We evaluate whether a workflow produces the requested results and whether its execution respects the task's temporal requirements. These assessments are complementary: a program can match the reference output while using information released after the required cutoff or selecting an incorrect historical state.
Result evaluation compares tables, figures, and text with the task requirements and hidden reference outputs. Human workflow auditing examines the program and execution records for PIT violations and other research errors. 
The following subsections describe result evaluation, human auditing, audit
coverage and quality control, and reporting metrics.

\hide{
Evaluation follows the workflow from the research request to execution and submission, then examines two aspects of the same submission. Result evaluation compares its outputs with the hidden reference under the output contract. Human workflow auditing uses a 20-category multilabel error codebook, with a particular focus on PIT validity. For PIT assessment, auditors use the task's temporal rules, field semantics, program, and execution traces to examine how information enters the analysis. Result judgments and audit labels are retained separately and linked through their shared submission version. Figure~\ref{fig:evaluation-protocol} summarizes the protocol.
}


\subsection{Result Evaluation}

The evaluator in the final release compares tables programmatically and assesses images and text with LLM judges using task-specific checklists.

\paragraph{Tables.}
For tasks with a fixed table format, we compare the expected structure and the values at corresponding positions. Column names and numeric values are normalized before comparison, with task-specified precision and units respected. Row multiplicity is preserved, and ordering is enforced only when required by the task.

For tasks with flexible formatting, semantically equivalent outputs may differ in column names, order, or auxiliary columns. We align fields using accepted names and aliases, permit column reordering, and apply declared task rules for auxiliary columns and alternative layouts. 

\paragraph{Images and text.}
LLM judges score images and text against checklists that are either supplied or
generated by a model. Each checklist breaks the task's output requirements into
individual items for assessment. 

\paragraph{Mixed outputs and passing criteria.}
For tasks with multiple output types, we aggregate their scores using relative weights of $3:2:1$ for tables, images, and text, respectively, reflecting their typical importance in the task output. Image and mixed-output tasks pass when all mandatory items pass and judging is available; the aggregate score summarizes item-level results.

\subsection{Human Workflow Audit}
\label{sec:pit_audit}

Auditors review the task instructions, generated program, and submitted and reference outputs together, using execution traces to identify errors and locate their causes. Each finding records the error type, the violated requirement, the relevant data source or program step, and supporting evidence. We accept equivalent implementations, including temporal filtering in SQL or Python before records affect a selection or calculation. A PIT violation requires evidence of a violated task rule in executed steps contributing to a required decision or output; matching the reference output does not rule out such a violation.

We classify errors into 20 types across seven groups: (A) temporal and condition filtering, (B) table and field interpretation, (C) data analysis and planning, (D) joins, (E) output requirements, (F) agent behavior and context, and (G) execution and SQL dialect errors. Appendix \ref{app:audit_codebook} gives the complete definitions. Each error receives the most specific applicable label, and a program may receive multiple labels for distinct errors. We focus particularly on three types of PIT violation:

\begin{itemize}[leftmargin=*]
    \item \textbf{A1: Temporal-semantic misuse.} Confusing business-valid time, information availability, and system-recording time. For example, a program uses an annual report for a December decision because its reporting period ends in December, although the report is disclosed the following March.
    \item \textbf{A2: Current state substituted for history.} Using current classifications, ratings, or status in place of those at the research date. For example, constructing a historical stock universe from current listing status can exclude companies that were listed then but have since delisted.
    \item \textbf{A3: Faulty historical reconstruction.} Using appropriate historical sources and correctly interpreting their dates, but incorrectly handling validity periods, record versions, joins, or duplicate removal. For example, a program checks only the start date of a historical industry classification and retains a record that had already expired by the research date.
\end{itemize}

\paragraph{Audit coverage and quality control.}
We audit 900 model submissions on 100 tasks spanning the four asset domains of the 408-task BuySideFlow benchmark. 
GPT-6.1 SOL reviewed each audited program and its execution trace using the error definitions in Appendix~\ref{app:audit_codebook}, proposing labels with
supporting evidence. 
An agent also assists with evidence retrieval, and
domain experts verify the evidence and confirm
the final audit labels. A financial data scientist independently annotated 200 submissions, two per audited task, with the primary audit labels withheld. We compare these annotations with the revised primary audit labels and report raw agreement and Cohen's $\kappa$ for each error type, group, and the presence of any PIT violation. Confidence intervals use 10,000 bootstrap resamples of task identifiers, keeping submissions from the same task together. Unresolved judgments are excluded only from the affected comparisons. Appendix~\ref{app:audit_agreement} details the comparison and its revision status.
All reported audit counts
and error frequencies are based on the decisions of a senior meta-reviewer.

\paragraph{Static PIT screening.}
An auxiliary rule-based checker extracts information cutoffs and historical-validity dates from task instructions, then matches predefined table/field uses and temporal predicates in submitted SQL/Python. It emits diagnostic flags for predefined temporal-rule patterns, plus review candidates for suspected future-label use. These signals are separate from result scores. Appendix~\ref{app:static_pit} specifies the rules and comparison protocol.

\subsection{Evaluation Metrics}
\label{sec:eval_metrics}

We report two complementary metrics to assess whether agents
produce correct results and whether their workflows satisfy
point-in-time (PIT) requirements. 

\paragraph{Result Accuracy.}
Let $\mathcal{E}$ denote all evaluated runs, and let
$r_i \in \{0,1\}$ indicate whether run $i$ passes the
result evaluation. Result Accuracy (RA) is defined as
\begin{equation}
    \mathrm{RA} =
    \frac{\sum_{i \in \mathcal{E}} r_i}{|\mathcal{E}|}.
\end{equation}
Each agent configuration is evaluated on all 408 tasks.
Runs that fail to produce a passing output are counted
as unsuccessful.

\paragraph{PIT Violation Rate.}
To examine temporal errors that may remain undetected by
result evaluation, we measure the proportion of
result-passing runs with confirmed PIT violations.
Let $\mathcal{A}^{+}$ denote the audited result-passing
runs, and let $v_i = 1$ if run $i$ has at least one
confirmed PIT violation (A1--A3), and $0$ otherwise.
The pass-side PIT Violation Rate is
\begin{equation}
    \mathrm{PVR}_{\mathrm{pass}} =
    \frac{\sum_{i \in \mathcal{A}^{+}} v_i}
    {|\mathcal{A}^{+}|}.
\end{equation}
Each run is counted once, even if multiple PIT violations
are identified.

\paragraph{Reporting.}
Each run is a configuration--task pair. We report $\mathrm{PVR}_{\mathrm{pass}}$ with its numerator and denominator for all result-pass records in the shared 100-task audit. Records with confirmed violations enter the numerator despite other unresolved issues; unresolved-only records remain in the denominator and are reported separately, alongside unaudited records. Absence of confirmed violations does not establish PIT compliance. The rate does not measure output impact or the fraction of all violating programs missed by result checks. Appendix~\ref{app:output_impact} separately examines output impact under targeted data perturbations.

Error frequencies and co-occurrences are reported separately for audited pass and fail records. Categories may overlap; these groups are not pooled to estimate PIT prevalence across all runs.

\section{Results and Analysis}
\label{sec:results}

\begin{table*}[t]
\centering
\caption{Result accuracy (\%) overall, by asset domain, and by reference workflow for nine baseline configurations. Parentheses give task counts; Overall pools all 408 tasks. Bold indicates column maxima, including ties.}
\label{tab:model-audit-results}
\small
\setlength{\tabcolsep}{3pt}
\setlength{\extrarowheight}{1.5pt}
\begin{tabular*}{\textwidth}{@{\extracolsep{\fill}}llrrrrrrr@{}}
\toprule
 & & & \multicolumn{4}{c}{Asset domain} & \multicolumn{2}{c}{Reference workflow} \\
\cmidrule(lr){4-7}\cmidrule(l){8-9}
Agent & Model & Overall & Equity & Fund & \shortstack{Fixed\\income} & Macro & SQL-native & SQL--Python \\
 &  & (408) & (166) & (125) & (93) & (24) & (254) & (154) \\
\midrule
\multirow{5}{*}{SWE-agent} & GPT-5.5 & 35.0 & \textbf{34.9} & 32.0 & 29.0 & 75.0 & 38.2 & \textbf{29.9} \\
 & DeepSeek-V4-Flash & 30.9 & 25.3 & 30.4 & 29.0 & \textbf{79.2} & 33.1 & 27.3 \\
 & Claude-Opus-4.7 & 29.9 & 28.3 & 23.2 & 31.2 & 70.8 & 32.7 & 25.3 \\
 & GLM-5.1 & 19.6 & 16.9 & 12.0 & 23.7 & 62.5 & 24.4 & 11.7 \\
 & Qwen3.6-Plus & 16.2 & 13.9 & 13.6 & 17.2 & 41.7 & 19.7 & 10.4 \\
\addlinespace[2pt]
Claude Code & Claude-Opus-4.7 & 24.0 & 22.3 & 18.4 & 23.7 & 66.7 & 25.2 & 22.1 \\
\addlinespace[2pt]
Codex & GPT-5.5 & \textbf{36.8} & 30.7 & \textbf{40.8} & \textbf{33.3} & 70.8 & \textbf{42.1} & 27.9 \\
\addlinespace[2pt]
\multirow{2}{*}{Spider-Agent-Lite} & GPT-5.5 & 28.7 & 27.1 & 26.4 & 21.5 & \textbf{79.2} & 31.9 & 23.4 \\
 & DeepSeek-V4-Flash & 22.3 & 19.9 & 14.4 & 24.7 & 70.8 & 23.2 & 20.8 \\
\bottomrule
\end{tabular*}
\end{table*}

\subsection{Experimental Setup and Audit Coverage}
\label{sec:experimental_setup}

We evaluate nine agent configurations. The adapted SWE-agent baselines \citep{yang2024sweagent} use GPT-5.5, DeepSeek-V4-Flash, Claude-Opus-4.7, GLM-5.1, and Qwen3.6-Plus. Additional baselines use Claude Code with Claude-Opus-4.7, Codex with GPT-5.5, and Spider-Agent-Lite with GPT-5.5 or DeepSeek-V4-Flash. Each LLM generates and revises SQL/Python using schema and execution tools. SWE-agent dispatches actions, returns feedback, enforces run limits, and saves trajectories (Figure~\ref{fig:evaluation-protocol}a). Each baseline configuration contributes one final run record for each of the 408 tasks, giving 3,672 records including runs without a final submission. A separate SWE-agent--GPT-5.5 configuration, denoted \emph{No PIT guidance}, contributes 408 further records and is reported separately from the baselines.

The error analysis uses 900 configuration--task records covering a shared set of 100 tasks across the nine baseline configurations. Of these records, 380 pass result evaluation and 520 fail. Run settings are documented in Appendix~\ref{app:run_settings}.

\subsection{Research Task Completion}
\label{sec:task_completion}

The final release has 993/3,672 passing runs (27.0\%), with full-match accuracy ranging from 16.18\% to 36.76\%. For 169 tasks (41.4\%), none of the nine configurations passes; 10 tasks (2.5\%) are passed by all nine. Table~\ref{tab:model-audit-results} reports accuracy overall, by asset domain, and by reference workflow for all nine configurations.

Macroeconomic tasks have the highest result accuracy within each configuration, with 68.5\% across their 216 runs. These contrasts describe performance on different task mixtures; they do not isolate the effect of asset domain.

\subsection{Error Analysis}
\label{sec:temporal_judgments}

\begin{figure*}[t]
  \centering
  \includegraphics[width=\textwidth]{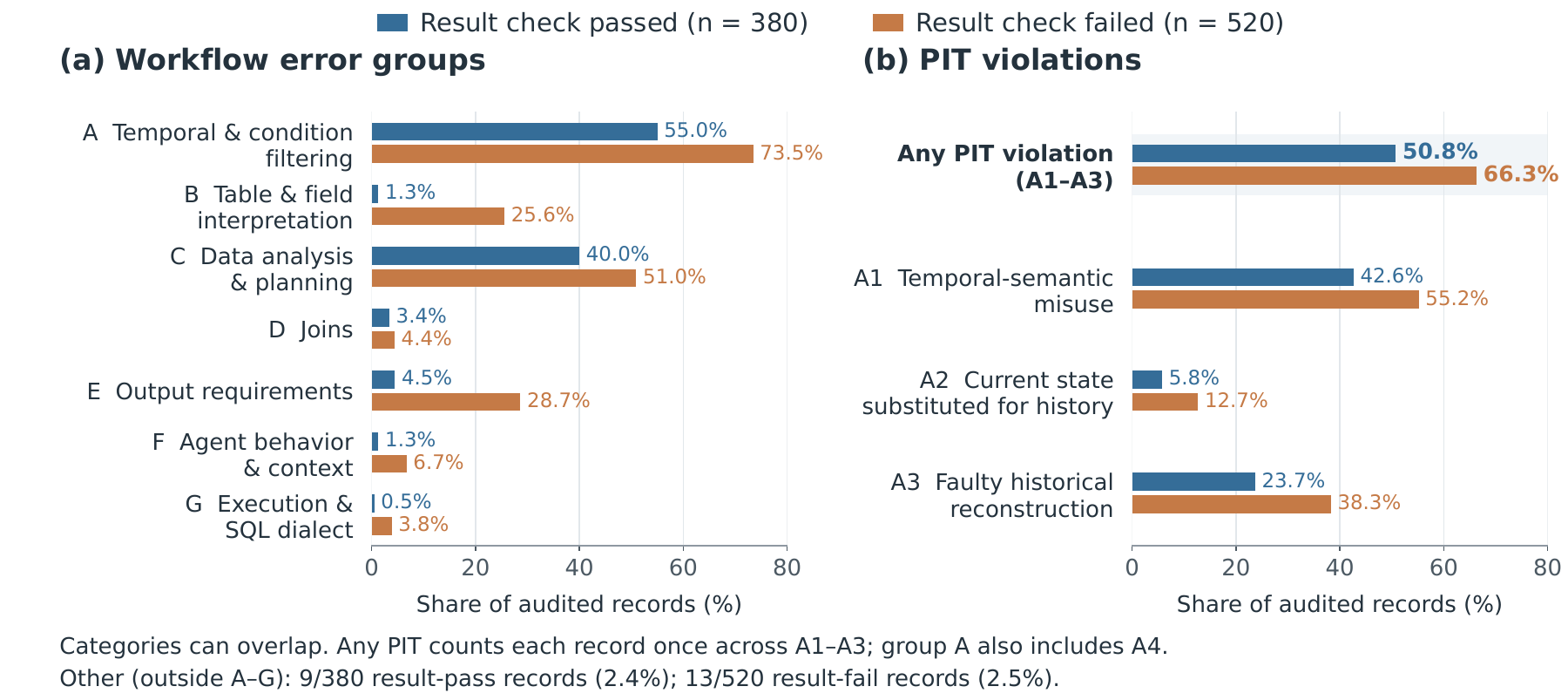}
  \caption{Confirmed error frequencies among audited records: (a) seven error groups; (b) any PIT violation and PIT types A1--A3. Each record is counted at most once per group or type. Categories can overlap.}
  \label{fig:error_distribution}
\end{figure*}

PIT violations (A1--A3) occur in 193 of 380 audited result-pass records (50.8\%) and 345 of 520 result-fail records (66.3\%). Across the seven error groups in Figure~\ref{fig:error_distribution}a, temporal and condition filtering (A) is the most frequent in both sets, followed by data analysis and planning (C). Temporal-semantic misuse (A1) is the most frequent PIT type in both sets (Figure~\ref{fig:error_distribution}b). Task/reference issues are also flagged in 69 result-fail records and can coexist with confirmed model errors.

Independent annotations agree with the revised primary audit on any PIT violation for 182/200 submissions (91.0\%, Cohen's $\kappa=0.818$, 95\% CI: 0.720--0.900). Among the 83 result-pass submissions, agreement is 78/83 (94.0\%, $\kappa=0.878$, 95\% CI: 0.762--0.975). Agreement for all 20 error types and seven groups is reported in Appendix~\ref{app:audit_agreement}. The PIT-labeled pass count depends on how the two disputed listing conditions are classified: treating both as non-PIT gives 188/380 (49.5\%), while treating both as PIT and adding the identified unannotated cases gives 198/380 (52.1\%; Appendix~\ref{app:output_impact}).

Among the 193 result-pass records with PIT violations, targeted single-event perturbations exposed non-rounding discrepancies in 111 and execution failures in two more (113/193, 58.5\%). This denominator includes 58 records assessed without perturbation execution (Appendix~\ref{app:output_impact}).

Combined diagnostic and review flags mark 200 of 883 nonempty programs, of which 188 contain PIT violations according to the revised primary audit (94.0\% precision, 34.9\% recall; Appendix~\ref{app:static_pit}). They flag 49/193 result-pass PIT violations (25.4\% recall), leaving 144 unflagged.

\subsection{Where Temporal Requirements Fail in Execution}
\label{sec:cases}

\paragraph{Version-selection order.}
Task \texttt{fund\_110} asks for funds in a specified category as of 16 February 2026 and requires selecting the applicable classification version before testing its category. DeepSeek-V4-Flash instead filters for the target category before ranking classification versions. This order can retain an older matching classification when the applicable version belongs to a different category. The submission matches the reference output but violates the task's prescribed selection order (A3). The evidence identifies the ordering violation, without identifying a particular obsolete classification in the output.

\paragraph{Information availability.}
To identify main funds managed by a specified manager as of 24 December 2025, task \texttt{fund\_041} requires both an active appointment and disclosure by that date. DeepSeek-V4-Flash retrieves the disclosure-date field and records showing an appointment effective on 23 December but disclosed on 25 December. These records remain in context at submission, yet the final query checks only appointment intervals and includes the fund. The agent corrects an unknown-column error after execution feedback but leaves the disclosure cutoff unimplemented. This result-failing submission provides record-level evidence of an A1 violation: the needed temporal information was retrieved but not enforced.

\paragraph{Historical-state selection.}
Task \texttt{stock\_002} requires selecting Chinese machinery-sector stocks using their listing state as of 31 October 2023. Qwen3.6-Plus instead filters on the current listing status stored in the database. This substitutes current state for the required historical state (A2) and can exclude stocks that were listed at the research date but subsequently delisted.

Appendix~\ref{app:temporal_fields} describes their schema evidence, and Appendix~\ref{app:construction_audit_details} retains additional defects and the limits of output-impact attribution.

Appendix~\ref{sec:pit_guidance_comparison} presents the PIT guidance comparison.

\section{Related Work}
\label{sec:related_limits}
\label{sec:related_work}

\paragraph{Financial databases and analytical workflows.}
BULL provides financial NL--SQL pairs over fund, stock, and macroeconomic databases, while FINCH combines it with other financial SQL resources \citep{zhang2024finsql,singh2025finch}. Spider~2.0 evaluates enterprise SQL/Python/dbt workflows \citep{lei2025spider2}. DAB exposes SQL and Python tools for analysis over real datasets with domain rules and data-quality issues, using deterministic answer checks \citep{ma2026dab}.

\paragraph{Financial research and quantitative tasks.}
FinGAIA combines heterogeneous tools with financial tasks, including historical holdings, document versions, and Python calculations \citep{zeng2025fingaia}. FinSkillBench evaluates investment subtasks with PIT data tools and deterministic checks on structured outputs. Expert Python skills are optional, and its separate nine-stage composition is a single-model, single-mandate probe \citep{bek2026finskillbench}.

\paragraph{Point-in-time controls and audits.}
AlphaQT-Bench gates functional accuracy of generated factors on prefix-causality checks for dependence on future observations \citep{luo2026alphaqt}. FinanceGym enforces publication cutoffs through retrieval and invalidates runs with detected leakage \citep{xiao2026financeharness}. OpenPM gates portfolio-agent inputs by availability, includes historical data vintages, and reports PASS/FAIL/UNKNOWN contamination certificates alongside performance \citep{cai2026openpm}. PIT Discovery separately measures retrieval quality and temporal violations in selected interaction-memory dependencies \citep{zhou2026historical}. BuySideFlow audits executed dependencies in agent-written SQL and SQL--Python workflows for Chinese buy-side research against task-specific disclosure cutoffs, historical-state requirements, and record-version rules. Table~\ref{tab:benchmark-comparison} provides an extended comparison.

\section{Conclusion}
\label{sec:conclusion}

BuySideFlow evaluates generated SQL and SQL--Python workflows against the disclosure, historical-state, and version-selection requirements of Chinese buy-side research. An expert audit of 900 records from nine configurations on a shared 100-task subset identifies at least one PIT violation in 193 of the 380 records that pass result checks (50.8\%). Output agreement therefore does not establish that a workflow respects the task's temporal requirements.

\section*{Limitations}

\paragraph{Data scope.}
BuySideFlow uses Chinese buy-side tasks and a sample snapshot of one vendor's production database, limiting transfer to other markets and sources. Licensed read-only access for reproduction is described in Appendix~\ref{app:construction_audit_details}.

\paragraph{Runs and sampling scope.}
One retained run per model and task does not estimate generation variability. Models also pass different task subsets, so $\mathrm{PVR}_{\mathrm{pass}}$ comparisons reflect both performance and the composition of each passing set. We have not quantified how task construction affects the reported statistics.

\paragraph{Audit reliability.}
PIT judgments depend on interpreting task rules, field meanings, and program evidence. Independent review covers 200 of the 900 audited submissions. Agreement with the revised primary audit varies across error types, and rare categories have few positive examples (Appendix~\ref{app:audit_agreement}). Two listing-eligibility conditions were coded inconsistently. Appendix~\ref{app:output_impact} reports sensitivity to alternative coding.

\paragraph{Reference conventions and output impact.}
Reference conventions and underspecified task conditions can affect result judgments. Perturbation runs on 39 tasks exposed three reference programs with temporal-selection defects whose corrections left the snapshot answers unchanged. References for the remaining 369 tasks underwent snapshot validation and targeted code checks but were not exercised in this perturbation study. The output-impact analysis establishes discrepancies under targeted conditions, not their frequency in practice (Appendix~\ref{app:output_impact}). Some cases depend on study-specific reference variants or event placements without entity-level precedent. We have not isolated PIT violations' contribution to result failure. Look-ahead bias from model pretraining is outside the audit's scope.

\bibliography{references}

\clearpage
\appendix
\section{Temporal Semantics and Case Evidence}
\label{app:temporal_fields}

Table~\ref{tab:temporal-field-checklist} maps the case studies to historical sources and temporal fields under the requirements in Section~\ref{sec:pit_requirements}. Business validity, public disclosure, and database recording have distinct meanings; any recording-time proxy follows the applicable task rule.

\begin{table*}[t]
\centering
\small
\setlength{\tabcolsep}{4pt}
\renewcommand{\arraystretch}{1.08}
\caption{Temporal sources for the case studies. The final column describes their task-specific use and temporal semantics.}\label{tab:temporal-field-checklist}
\begin{tabular}{@{}>{\raggedright\arraybackslash}p{0.25\textwidth}>{\raggedright\arraybackslash}p{0.34\textwidth}>{\raggedright\arraybackslash}p{\dimexpr0.41\textwidth-16pt\relax}@{}}
\toprule
Case and data dependency & Historical source and time fields & Interpretation and evidence boundary \\
\midrule
\texttt{stock\_002}: listing state\newline
\path{secumain.LISTEDSTATE}
& \path{lc_liststatus}, joined on \path{INNERCODE}.\newline
\path{CHANGEDATE}, \path{CHANGETYPE}, \path{INSERTTIME}.
& Reconstruct the state at the research date from applicable changes. This requires an initial state, event-to-state mapping, same-day ordering, and market coverage. \path{CHANGEDATE} is not automatically a public-disclosure date.\\[5pt]

\texttt{stock\_002}: industry\newline
\path{lc_exgindustry.IFPERFORMED}
& Same-table history by \path{COMPANYCODE}, \path{STANDARD}, and classification level.\newline
\path{INFOPUBLDATE}, \path{CANCELDATE}.
& The schema describes the flag as the latest valid classification. Historical selection follows the task's interval convention, without an additional current-validity filter. Using publication as the interval start follows that convention, not a general equivalence between disclosure and business validity.\\[5pt]

\texttt{fund\_041}: manager tenure\newline
\path{mf_fundmanagernew}
& Same-table tenures by \path{INNERCODE}, \path{PERSONALCODE}, and \path{POSTNAME}.\newline
\path{ACCESSIONDATE}, \path{DIMISSIONDATE}, \path{INFOPUBLDATE}.
& Check both the appointment interval and disclosure cutoff. Preserve co-managers, distinct posts, and repeat tenures. The table records historical managers and assistants.\\[5pt]

\texttt{fund\_110}: fund classification\newline
\path{mf_jyfundtype}
& Same-table classifications with \path{INNERCODE}, \path{TRANSCODE}, and \path{STANDARD}.\newline
\path{EFFECTIVEDATE}, \path{CANCELDATE}, \path{INSERTTIME}.
& Under this task's rule, select the applicable version before testing its category. Identifier correspondence and version grain require care. The table has no dedicated public-disclosure field, so neither effective time nor insertion time automatically establishes public availability.\\
\bottomrule
\end{tabular}
\end{table*}

\paragraph{Evidence boundary.}
The field mapping uses table names, column definitions, and comments in \path{gildata_schema.json}. Task-level data validation is summarized in Appendix~\ref{app:construction_audit_details}. The grouping fields listed here identify candidate dimensions for interpreting sources, not independently validated alternative reconstruction queries.

Reconstructing earlier database states requires retained record versions and their replacement or withdrawal history. An insertion or last-update timestamp alone does not establish that history.

The complete 27-entry schema-level inventory, including source-line locations, other candidate paths, and unresolved items, is retained separately in \path{temporal_field_inventory.md}. It is a candidate checklist, not a count of validated reconstructions.

\subsection{Worked Example: \texttt{stock\_069}}
\label{app:stock069_workflow}

Table~\ref{tab:stock069-workflow} details the reference operations and submission evidence for the successive-date industry-weight example in Section~\ref{sec:pit_requirements}.

\begin{table}[H]
\centering
\small
\setlength{\tabcolsep}{4pt}
\renewcommand{\arraystretch}{1.1}
\caption{The recurring historical-state requirement in \texttt{stock\_069}. Reference operations illustrate the task; the last column identifies evidence to inspect in a submitted workflow.}
\label{tab:stock069-workflow}
\begin{tabular}{@{}>{\raggedright\arraybackslash}p{0.19\linewidth}>{\raggedright\arraybackslash}p{0.42\linewidth}>{\raggedright\arraybackslash}p{\dimexpr0.39\linewidth-16pt\relax}@{}}
\toprule
Research requirement & Reference operation & Submission evidence \\
\midrule
Constituents at each date & Read members and weights from each dated index-weight snapshot & Dates and member keys used to select the universe \\
\addlinespace[3pt]
Industry applicable at $d$ & Restrict disclosure and cancellation dates at $d$; apply the effective-date bound required by the source & Historical source, date boundaries, and records retained for each company and date \\
\addlinespace[3pt]
Applicable version & Rank eligible records within each company--date group using the source's version ordering & Partition keys and the order of temporal filtering and ranking \\
\addlinespace[3pt]
Industry exposure at $d$ & Join on company and date, sum weights by industry, and rank within each date & Join keys and the industry assignments contributing to each aggregate \\
\bottomrule
\end{tabular}
\end{table}

\section{Error Audit Codebook}
\label{app:audit_codebook}

Table~\ref{tab:audit-codebook} defines the 20 second-level error codes used in the audit. A1--A3 constitute PIT violations. A4 and C2 cover non-PIT filtering and time-series errors, respectively. The evidence requirements follow Section~\ref{sec:protocol}. \texttt{Task/reference issue} and \texttt{Infrastructure/runtime} are separate markers, not additional error categories.

Each defect receives the most specific applicable code. Multiple codes require distinct defects or separately evidenced process failures. A demonstrated violation of an explicit task rule can be confirmed even if the output passes evaluation, provided that the violating dependency lies on an executed path to the submitted output or required analytical decision. Unexecuted branches and unused intermediates do not suffice. The effect on the current result is a separate question. Unsettled requirements or indeterminate program behavior remain unresolved or are recorded as task/reference issues. A downstream exception or missing artifact alone does not establish an additional error. Output-contract codes require an independent defect in the relevant output component, not correctness of the entire program.

\begin{table*}[t]
\centering
\small
\setlength{\tabcolsep}{4pt}
\renewcommand{\arraystretch}{1.08}
\caption{Multilabel error taxonomy. Coding boundaries distinguish the underlying defect from its downstream symptoms.}\label{tab:audit-codebook}
\begin{tabular}{@{}>{\raggedright\arraybackslash}p{0.065\textwidth}>{\raggedright\arraybackslash}p{0.24\textwidth}>{\raggedright\arraybackslash}p{\dimexpr0.695\textwidth-16pt\relax}@{}}
\toprule
Code & Error type & Definition and coding boundary \\
\midrule
\multicolumn{3}{@{}l}{\textbf{A. Temporal and condition filtering}}\\*[3pt]
A1 & Temporal-semantic misuse &
Confuses business-valid time, system-recording or ingestion time, disclosure availability, or the observation cutoff. For example, a reporting period is treated as evidence of availability. Interpretation follows the task and field semantics.\\[5pt]
A2 & Current state substituted for history &
Uses current constituents, classifications, ratings, or status snapshots in place of the state valid at the research time. The same substitution is not also coded as faulty historical reconstruction.\\[5pt]
A3 & Faulty historical reconstruction &
Uses the correct historical source and time semantics but implements validity intervals, version selection, historical joins, or reconstruction deduplication incorrectly. The same historical-join defect is not also D1.\\[5pt]
A4 & Task conditions and sample selection &
Omits, adds, or misapplies a non-PIT eligibility or selection rule specified by the task, inputs, or mandatory rules actually supplied to the agent. Covers market, board, security type, rating, size thresholds, designated samples, comparisons, Boolean logic, quantifiers, NULL handling, and Top-$K$ or tie retention. Field meaning belongs to B2, time-series operations to C2, numeric metric or rank computation to C3, and joins to D1.\\[6pt]

\multicolumn{3}{@{}l}{\textbf{B. Schema linking}}\\*[3pt]
B1 & Table linking &
Selects an incorrect table, omits a required table, or invents one. Current-state substitution and temporal-semantic misuse take A2 and A1, respectively.\\[5pt]
B2 & Column linking and field semantics &
Misidentifies, omits, or invents a field, or misinterprets its encoding, units, or join-key identity. Correctly understood keys used in a faulty join belong to D1. PIT time-semantic errors belong to A1.\\[6pt]
\bottomrule
\end{tabular}
\end{table*}

\begin{table*}[t]
\centering
\small
\setlength{\tabcolsep}{4pt}
\renewcommand{\arraystretch}{1.08}
\ContinuedFloat
\caption{Multilabel error taxonomy. Coding boundaries distinguish the underlying defect from its downstream symptoms.}
\begin{tabular}{@{}>{\raggedright\arraybackslash}p{0.065\textwidth}>{\raggedright\arraybackslash}p{0.24\textwidth}>{\raggedright\arraybackslash}p{\dimexpr0.695\textwidth-16pt\relax}@{}}
\toprule
Code & Error type & Definition and coding boundary \\
\midrule
\multicolumn{3}{@{}l}{\textbf{C. Data analysis and query planning}}\\*[3pt]
C1 & Multi-step planning &
Omits a necessary subtask or plans dependencies or subresult composition incorrectly. Requires direct evidence of a planning defect. Task complexity, incompletion, or a local coding error alone is insufficient.\\[5pt]
C2 & Time-series and window implementation &
Implements sorting, lags, rolling windows, resampling, or calendar alignment incorrectly without using unavailable information or misrepresenting historical states or versions. Future-information use, current-state substitution, and faulty historical reconstruction belong to A1, A2, and A3. C2 is co-assigned only for a distinct time-series computation defect. Financial formulas belong to C3.\\[5pt]
C3 & Business metrics and numeric computation &
Miscomputes financial formulas, non-temporal aggregates, statistics, or missing-value handling, including scaling, currency conversion, and premature rounding during computation. Misunderstood field meaning belongs to B2. Incorrect final presentation of a correctly computed value belongs to E2.\\[6pt]

\multicolumn{3}{@{}l}{\textbf{D. Joins}}\\*[3pt]
D1 & Join implementation &
Understands fields and keys correctly but uses an incorrect SQL join or Pandas merge type, predicate, cardinality, or join deduplication. Key-identity errors belong to B2, and historical-reconstruction join errors to A3.\\[6pt]

\multicolumn{3}{@{}l}{\textbf{E. Output contracts}}\\*[3pt]
E1 & Tabular structure and submission &
Violates the contract for column names or order, result variables, non-chart output files, serialization, or submission. An absent artifact caused solely by unfinished computation does not establish E1.\\[5pt]
E2 & Display precision and units &
Computes the relevant value and base scale correctly but displays the wrong unit, percentage sign, or rounding precision. A computational conversion error belongs to B2 or C3, without an additional E2 for the same defect.\\[5pt]
E3 & Output labels and enumerations &
Determines the relevant business category correctly but renders a string or enumeration that violates the output contract. Incorrect classification rules, thresholds, or field meanings instead follow the corresponding A, B, or C code.\\[5pt]
E4 & Chart specifications &
Violates an explicit requirement for chart presence, type, axes, series, legend, or file. Unfinished computation alone does not establish a missing-chart defect. Incorrect source data are coded by their underlying mechanism. Aesthetic preferences are not criteria.\\[6pt]
\bottomrule
\end{tabular}
\end{table*}

\begin{table*}[t]
\centering
\small
\setlength{\tabcolsep}{4pt}
\renewcommand{\arraystretch}{1.08}
\ContinuedFloat
\caption{Multilabel error taxonomy. Coding boundaries distinguish the underlying defect from its downstream symptoms.}
\begin{tabular}{@{}>{\raggedright\arraybackslash}p{0.065\textwidth}>{\raggedright\arraybackslash}p{0.24\textwidth}>{\raggedright\arraybackslash}p{\dimexpr0.695\textwidth-16pt\relax}@{}}
\toprule
Code & Error type & Definition and coding boundary \\
\midrule
\multicolumn{3}{@{}l}{\textbf{F. Agent and context}}\\*[3pt]
F1 & Budget exhaustion &
A documented step, token, time, or cost limit causes termination or incompletion. May coexist with an earlier error. A missing submission or truncated trace alone is not evidence of a limit being reached.\\[5pt]
F2 & Hallucinated environment capability &
Relies on a nonexistent file, tool, package, or runtime capability and leaves the issue unresolved. Invented tables or columns belong to B1 or B2. Actual service failures receive the separate \texttt{Infrastructure/runtime} marker.\\[5pt]
F3 & Ineffective repair loop &
Repeated ineffective attempts despite usable feedback leave an unresolved problem that causes incompletion. May coexist with the original defect or F1 when each has supporting evidence. Ordinary debugging that succeeds is excluded.\\[5pt]
F4 & Unsupported early stopping &
Actively stops or abandons the task despite necessary remaining steps and no justified stopping condition. Non-budget termination alone is insufficient. A limit-triggered stop belongs to F1. Justified stopping at an established task or environment barrier is excluded. A correct artifact submitted incorrectly belongs to E1.\\[6pt]

\multicolumn{3}{@{}l}{\textbf{G. Execution and dialect}}\\*[3pt]
G1 & General syntax or runtime implementation &
Leaves a syntax, type, indexing, or API-use error unresolved when no more specific mechanism explains it. An exception directly caused by a table, column, computation, or dialect defect does not also receive G1.\\[5pt]
G3 & SQL dialect incompatibility &
Uses syntax or functions from another SQL dialect that the target MySQL engine does not support or interprets differently. Requires evidence of a dialect mismatch. Ordinary typos and Pandas API errors belong to G1 or a more specific code.\\
\bottomrule
\end{tabular}
\end{table*}

\clearpage
\section{Construction and Audit Details}
\label{app:construction_audit_details}

\subsection{Task Packages and Database Evidence}

\paragraph{Task metadata and environment.}
Thirty-two tasks specify \texttt{inputs.as\_of\_date}; others express time anchors in instructions or task-specific inputs. All evaluations use the fixed \texttt{Gildata} sample identified by \texttt{Gildata\_20260430\_v1}. The database maintainer confirms that its data have remained unchanged since 30 April 2026. We provide the 408 tasks, reference programs and answers, and background-knowledge files. Use of the database for benchmark construction and evaluation is authorized. The sample is hosted on Alibaba Cloud, with read-only credentials available by application under the existing license. References remain hidden from agents during evaluation.

\paragraph{Sample scope and retention.}
The shared sample exposes 110 tables and 4,547 columns. It retains data on both sides of individual research dates: the \texttt{stock\_002} validation extract contains 1,554 financial rows disclosed after its cutoff, and its historical candidates include subsequently delisted companies. Programs must apply the task's temporal selection. The validation queries record candidate keys, initial events, competing versions, and required periods within the retained sources. The available package does not include the original production-to-sample extraction manifest or complete per-table retention rules. Reproduction uses the identified sample through licensed access; rebuilding an identical sample from the production database requires those extraction rules from the maintainer.

\paragraph{Run settings.}
\label{app:run_settings}
The inspected per-task configurations set temperature to zero, leave \texttt{top\_p} unset, and disable parallel tool calls.

\paragraph{Temporal-requirement coverage.}
Table~\ref{tab:temporal-task-requirements} counts requirements across all 408 released task statements, structured inputs, and applicable business rules, independently of model outcomes. Disclosure cutoffs constrain information availability; historical states constrain date-valid eligibility and attributes. Version selection includes prescribed financial-report and classification versions, but not ordinary deduplication or selection of the latest reporting period alone. Decision sequences require separate information or state selection at successive dates, including historical reconstruction and signal formation; rolling descriptive statistics alone do not qualify. Ex-post labels are realized outcomes following an earlier selection, event, feature, or allocation, rather than hypothetical future cash flows. Labels overlap: 346 tasks have at least one, while 62 involve other controls, such as trading-day windows. Assistant-generated labels underwent a cross-domain consistency review. The annotation package retains each task, its labels and rationale, source versions, and annotation date.

\paragraph{Task-level data validation.}
Construction review examined task statements, reference programs, and outputs together. The validation ledger records four evidence targets: candidate-universe coverage, initial states and subsequent changes, eligible record versions, and dates including required base periods. A reference returning an output establishes executability, while these checks examine the data needed for the calculation. The supplement contains data-level checks and separate recomputations for \texttt{stock\_002}, \texttt{stock\_069}, and \texttt{macro\_021}. It indexes all 408 tasks but does not establish complete historical coverage for every task. Section~\ref{app:stock002_construction} presents one checked path and its remaining coverage limits.

\paragraph{Construction and reference review.}
Practitioner review reduced the draft from 411 to 404 tasks; financial-definition review produced 408. The four rounds (Section~\ref{sec:construction}) have retained issue lists of 173, 50, 15, and 0 distinct tasks. Across rounds, 208 release tasks (51.0\%) had at least one recorded issue. Later rounds targeted unresolved issues and revisions, so these counts describe the review process rather than a fixed-sample error rate.

\subsection{Worked Construction Example: \texttt{stock\_002}}
\label{app:stock002_construction}

\paragraph{Research request and specification.}
The research objective is to identify machinery companies with accelerating quarterly revenue growth. The task fixes the universe to Shanghai/Shenzhen main-board A-shares listed at the end of 31 October 2023. It requires consolidated, single-quarter revenue, Q3 year-on-year growth above 20\%, and an increase over Q2 exceeding 10 percentage points. The output has six ordered columns: security code, name, Shenwan level-1 industry, Q2 growth, Q3 growth, and their difference. Filtering and sorting use unrounded values; display uses two decimals, with decreasing growth difference and then increasing security code.

\paragraph{Sources and historical selection.}
The reference combines the security master (\texttt{secumain}), listing events (\texttt{lc\_liststatus}), industry history (\texttt{lc\_exgindustry}), and revenue (\texttt{lc\_maindatanew}). End-of-day availability uses the exclusive boundary \texttt{2023-11-01}. Listing events must precede it in both change date and the permitted recording-time proxy; the latest eligible event determines listing status. Industry records must be disclosed and valid at the observation time. The applicable industry version is selected before testing machinery membership. The reproduction guide specifies join keys, status codes, and tie-breaking fields.

\paragraph{Versions and calculation.}
For each candidate, the reference reads the March, June, and September cumulative revenues in both 2022 and 2023. It filters by disclosure time before applying the background rules for report-version priority. Quarter-only report types are excluded from these cumulative inputs. Q2 revenue is half-year minus Q1 revenue; Q3 revenue is nine-month minus half-year revenue. Year-on-year growth is $100(R_{2023,q}/R_{2022,q}-1)$, using positive prior-year denominators and nonmissing current values. SQL performs selection and calculation, and the Python wrapper writes \texttt{result.csv}.

\paragraph{Validation evidence and acceptance.}
The retained-record checks enumerate 3,391 base candidates before listing and industry filters leave 239 machinery companies. All $239\times6=1,434$ required company--period pairs have eligible, nonmissing revenue values; 712 pairs contain competing pre-cutoff versions. The selected listing histories have matching initial listing dates and no same-day event ties in this extract. A separate Python/Decimal calculation from the extracted rows reproduces all 32 output rows, values, and ordering. For example, security \texttt{002529} has Q2 growth $-24.66\%$, Q3 growth $192.62\%$, and an increase of $217.28$ percentage points. The evidence files retain candidate decisions, selected record IDs, period coverage, and the output comparison.

Coverage remains conditional on the retained sources: 13 non-placeholder master records lack both a listing date and a pre-cutoff listing event, and no external reconciliation establishes that all master objects or overwritten versions are retained. The checks therefore support this reconstruction and calculation within the documented candidate range, with those coverage questions unresolved.

\paragraph{Task package and replay.}
The task package contains the instruction, a strict output contract, reference code, and saved CSV. Reference replay executes a temporary copy with saved outputs removed, then compares the regenerated result with the saved answer. The reproduction guide links construction decisions to source files, commands, and acceptance evidence.

\subsection{Result Evaluation Rules}
\label{app:result_evaluation_rules}

The current dataset contains 374 CSV tasks with \texttt{strict\_output\_schema} enabled, 27 CSV tasks without it, six image-oriented tasks, and one mixed task. Table~\ref{tab:result-evaluation-rules} documents the evaluator in the final release.

\begin{table*}[t]
\centering
\small
\setlength{\tabcolsep}{4pt}
\renewcommand{\arraystretch}{1.08}
\caption{Result-evaluation rules in the final release. Code entry points and normalization details are in the reproduction guide.}\label{tab:result-evaluation-rules}
\begin{tabular}{@{}p{0.19\linewidth}p{0.78\linewidth}@{}}
\toprule
Stage & Implemented rule \\
\midrule
Inputs and reference & Load task inputs, output contract, and reference artifacts. Saved outputs take priority; execute the reference when direct results are unavailable. \\
Output extraction & SQL returns a table. Python uses named result variables and captured figures. Unresolved extraction can trigger model-assisted recovery and re-execution. Successfully extracted outputs are retained even when their values mismatch. \\
Strict tables & Match required semantic fields, result components, and complete rows. Enforce column order after allowed presentation normalization, subject to declared task-specific comparison rules. \\
Non-strict tables & Permit column reordering. Accepted aliases, auxiliary columns, and alternative layouts follow declared task mappings. Field identities and complete rows remain required. Non-tabular outputs may invoke a model judge. \\
Numbers and units & Apply declared field precision; otherwise use recognized question precision, defaulting to four decimal places. Convert explicitly identified, dimension-compatible units without guessing scale factors. \\
Order and missing values & Apply task-specific sort rules or recognized ordering instructions; otherwise ignore row order. Preserve complete rows and multiplicity. Normalize actual missing values and blank cells. \\
Images and text & Apply supplied or model-derived checklists grounded in the question. Tables use programmatic comparison; images and text use model judges. Present categories use normalized table:image:text weights of $3:2:1$. \\
Passing and failures & Tables must match. Visual/mixed outputs require every mandatory item to pass and judging to be available; mandatory visual items require scores $\geq0.999$. Weighted scores summarize item results. Execution and comparison failures are retained. Manual result adjudications and PIT labels remain separate. \\
\bottomrule
\end{tabular}
\end{table*}

\subsection{Submission Audit Details}

\paragraph{Audit records and adjudication.}
Each audit record identifies the configuration and task, records the result-evaluation outcome, and links the supporting case evidence. Confirmed error labels and unresolved judgments or exceptions are stored separately; \texttt{None} denotes an empty set.

The expert assigned final labels after reviewing the AI proposals described in Section~\ref{sec:pit_audit}. This assisted review does not provide an independent inter-rater agreement estimate.

\paragraph{Independent annotation and agreement.}
\label{app:audit_agreement}
An independent financial data scientist annotated 200 submissions from the same 100 tasks, with primary audit labels withheld. Each comparison matches the task instructions and submitted SQL/Python program. The two replaced \texttt{bond\_001} submissions were independently annotated on their current 20-month versions. The primary \texttt{fund\_031} label was revised after annotation comparison; its independent label was retained. Table~\ref{tab:audit-agreement} therefore compares independent annotations with the revised primary audit.

A binary endpoint is positive when a relevant error is confirmed. If it has no confirmed error but a relevant unresolved judgment, that record is excluded for that endpoint. Group labels use the union of their constituent codes, and any PIT uses A1--A3. Task/reference and infrastructure markers are reported separately. This leaves 200 determinate any-PIT judgments; F1, G1, G3, and group G each have 197. Confidence intervals in Section~\ref{sec:temporal_judgments} are percentile intervals from 10,000 task-cluster bootstrap resamples (seed 20261012).

Across the 197 records with all 20 error types determinate, the complete label sets agree on 56 (28.4\%), with mean Jaccard similarity 0.610; empty--empty pairs receive a similarity of one. These set comparisons exclude Other and the separate exception markers. C1, E4, F2, and G3 have no positive annotations from either side, so their $\kappa$ is undefined. G1 has only two positive pairs. Agreement for these sparse types should be read together with their positive counts.

\begin{table*}[t]
\centering
\small
\setlength{\tabcolsep}{7pt}
\renewcommand{\arraystretch}{1.0}
\caption{Agreement between independent annotations and the revised primary audit on 200 submissions from 100 tasks. Positive counts are shown for both sources; $n$ excludes endpoint-specific unresolved judgments. Other and exception markers are outside the 20-code taxonomy. A dash indicates undefined $\kappa$.}
\label{tab:audit-agreement}
\begin{tabular}{lrrrrr}
\toprule
Endpoint & $n$ & \shortstack{Primary\\positive} & \shortstack{Independent\\positive} & \shortstack{Agreement\\(\%)} & $\kappa$ \\
\midrule
Any PIT & 200 & 119 & 105 & 91.0 & 0.818 \\
Any PIT, result pass & 83 & 39 & 34 & 94.0 & 0.878 \\
Any PIT, result fail & 117 & 80 & 71 & 88.9 & 0.759 \\
\midrule
A1 & 200 & 98 & 80 & 90.0 & 0.799 \\
A2 & 200 & 16 & 14 & 97.0 & 0.784 \\
A3 & 200 & 57 & 28 & 85.5 & 0.580 \\
A4 & 200 & 46 & 37 & 82.5 & 0.470 \\
B1 & 200 & 12 & 14 & 99.0 & 0.918 \\
B2 & 200 & 22 & 16 & 96.0 & 0.768 \\
C1 & 200 & 0 & 0 & 100.0 & -- \\
C2 & 200 & 48 & 27 & 83.5 & 0.468 \\
C3 & 200 & 83 & 63 & 81.0 & 0.594 \\
D1 & 200 & 11 & 4 & 96.5 & 0.519 \\
E1 & 200 & 38 & 73 & 80.5 & 0.532 \\
E2 & 200 & 0 & 9 & 95.5 & 0.000 \\
E3 & 200 & 1 & 0 & 99.5 & 0.000 \\
E4 & 200 & 0 & 0 & 100.0 & -- \\
F1 & 197 & 4 & 1 & 98.5 & 0.395 \\
F2 & 200 & 0 & 0 & 100.0 & -- \\
F3 & 200 & 8 & 3 & 97.5 & 0.535 \\
F4 & 200 & 0 & 1 & 99.5 & 0.000 \\
G1 & 197 & 2 & 2 & 100.0 & 1.000 \\
G3 & 197 & 0 & 0 & 100.0 & -- \\
\midrule
Group A & 200 & 132 & 118 & 90.0 & 0.788 \\
Group B & 200 & 33 & 30 & 95.5 & 0.831 \\
Group C & 200 & 102 & 77 & 78.5 & 0.572 \\
Group D & 200 & 11 & 4 & 96.5 & 0.519 \\
Group E & 200 & 39 & 80 & 76.5 & 0.465 \\
Group F & 200 & 9 & 5 & 97.0 & 0.557 \\
Group G & 197 & 2 & 2 & 100.0 & 1.000 \\
\midrule
Other & 200 & 3 & 1 & 99.0 & 0.496 \\
Task/reference issue & 200 & 13 & 1 & 94.0 & 0.135 \\
Infrastructure/runtime & 200 & 0 & 1 & 99.5 & 0.000 \\
\bottomrule
\end{tabular}
\end{table*}

\paragraph{Additional case evidence.}
The passing Claude-Opus-4.7 submission for \texttt{stock\_002} uses midnight upper bounds for industry and financial disclosures despite the run's explicit end-of-day cutoff. The inspected disclosure fields contain dates without intraday times, so these bounds retain the same records. The submission also checks only the first listing date. In a tested state where a selected company terminates its listing before the research date, it retains that company while the recomputed reference excludes it (Appendix~\ref{app:output_impact}).

In DeepSeek-V4-Flash's passing fund-size comparison (task \texttt{fund\_110}), a shared aggregation takes \texttt{MAX(COMBNVI)} for each fund and reporting period, covering 30 June and 31 December 2025. It omits the availability cutoff on \texttt{mf\_scaleanalysis.INSERTTIME}, which the schema retrieved at step 1 identifies as publication time. This omission is separate from the classification-ordering error. Multiple rows can represent classification dimensions, so the maximum alone does not establish use of a later revision. The effect on the selected funds and reported values has not been measured.

For \texttt{fund\_041}, the trace records schema retrieval at step 1, disclosure records at steps 3, 5, and 11, and correction of a column-name error between steps 6 and 8. A post-cutoff disclosure enters the final query output at step 10, but the evidence does not isolate PIT as the unique cause of result failure.

\paragraph{Error co-occurrences.}
Among records with confirmed PIT violations, 102 of 193 result passes and 301 of 345 result failures also have a confirmed non-PIT error. The remaining 91 passes and 44 failures have no confirmed non-PIT error; seven of these failures carry a task/reference issue. These counts describe error co-occurrences and do not isolate the effect of PIT violations on result failure.

\subsection{Static PIT Screening}
\label{app:static_pit}

\paragraph{Mechanism and inputs.}
The evaluated implementation in \texttt{time\_audit.py} uses lexical cues to distinguish historical-validity dates from information cutoffs. It recognizes predefined table and field names and matches SQL predicates and Python comparison patterns. Diagnostic rules cover runtime-date use, missing disclosure or validity guards, and patterns associated with current-state filtering. A separate heuristic flags suspected use of future outcome labels in selection or features. Screening uses task instructions and final submitted source text, without executing code or establishing end-to-end data dependencies. The workflow audit additionally examines execution traces.

\paragraph{Comparison protocol.}
Of the 900 audited records, 883 contain nonempty final programs. The 17 without submitted code have neither positive primary-audit PIT labels nor static flags and are excluded from the screening set. Table~\ref{tab:static-pit} reports diagnostic flags alone and their union with review candidates. For any PIT, a match means that a flagged program has at least one confirmed A1--A3 label. Precision is the fraction of flagged programs with such a label, and recall is the fraction of audit-positive programs flagged. The comparison package freezes the implementation, mapping, inputs, and record-level outputs; its manifest records their versions and freeze date.

For type-wise matching, runtime-date, financial-publication, risk-alert-information, and future-label flags map to A1. Current-state, listing-state, and historical-current-field flags map to A2. Risk-alert-effective, index-component, classification-vintage, and bond-concept-vintage flags map to A3. Among the 188 program-level matches under combined flags, 34 share no mapped type with the primary audit. Program-level precision therefore measures screening overlap rather than correctness of each specific alert.

\paragraph{Independent-label sensitivity.}
The independent-annotation subset contains 196 nonempty final programs, including 83 result passes. Four records without submitted code are excluded; one has an independently assigned PIT label. We compare the same available programs against each label source separately. Only 13 result-pass programs are flagged, making precision on this subset sensitive to a few label disagreements.

\begin{table*}[t]
\centering
\small
\setlength{\tabcolsep}{5pt}
\caption{Static PIT screening against audit labels. Combined flags include diagnostic flags and review candidates. $H$ counts audit-positive programs, $F$ flagged programs, and $T$ their intersection. Any-PIT rows use program-level overlap; A1--A3 rows require the mapped type to match. Precision and recall are percentages.}
\label{tab:static-pit}
\begin{tabular}{lrrrrrr}
\toprule
Comparison & $n$ & $H$ & $F$ & $T$ & Precision & Recall \\
\midrule
\multicolumn{7}{l}{\textit{Revised primary audit: screening coverage}} \\
Any PIT, diagnostic & 883 & 538 & 188 & 182 & 96.8 & 33.8 \\
Any PIT, combined & 883 & 538 & 200 & 188 & 94.0 & 34.9 \\
Result pass, diagnostic & 380 & 193 & 48 & 46 & 95.8 & 23.8 \\
Result pass, combined & 380 & 193 & 54 & 49 & 90.7 & 25.4 \\
\midrule
\multicolumn{7}{l}{\textit{Revised primary audit: type-wise matching, combined flags}} \\
A1 & 883 & 449 & 72 & 59 & 81.9 & 13.1 \\
A2 & 883 & 88 & 117 & 81 & 69.2 & 92.0 \\
A3 & 883 & 289 & 91 & 75 & 82.4 & 26.0 \\
\midrule
\multicolumn{7}{l}{\textit{Paired label comparison: any PIT, combined flags}} \\
Primary, all paired & 196 & 119 & 43 & 39 & 90.7 & 32.8 \\
Independent, all paired & 196 & 104 & 43 & 36 & 83.7 & 34.6 \\
Primary, paired result pass & 83 & 39 & 13 & 11 & 84.6 & 28.2 \\
Independent, paired result pass & 83 & 34 & 13 & 8 & 61.5 & 23.5 \\
\bottomrule
\end{tabular}
\end{table*}

\subsection{Output Sensitivity to Targeted Data Changes}
\label{app:output_impact}

\paragraph{Design.}
We examine the 193 result-pass records annotated with A1--A3 under changes to the retained data. After inspecting the implicated fields, we attempt execution of 135 submitted programs in 65 recorded scenarios covering 39 tasks. The remaining 58 records receive data and code inspection without perturbation execution. Each scenario represents one event and may change several associated fields or rows, such as a fund's termination dates and subsequent NAV records. Query rewriting exposes the same modified state to the submission and reference without changing the stored database. We compare their outputs on that state, first checking the unmodified state through the same rewriting path.

Event types follow patterns observed in the relevant tables, with placements chosen to exercise audited conditions. For example, 10,710 of 16,541 manager appointments effective from January 2024 through September 2025 were announced after taking effect. Among 20,462 NAV records dated 29 March 2024, 1,367 were first inserted on or after 1 April. These observations motivate delayed-disclosure and delayed-recording scenarios. The targeted placements do not sample their population frequencies. Two scenarios test unobserved classification-version overlap or intraday values in date-only fields and are excluded from impact classification. The 63 retained scenarios cover 133 records on the same 39 tasks, with one record's only retained run timing out.

\paragraph{Observed outcomes.}
For 113 records (58.5\% of 193), at least one scenario produces a discrepancy beyond the last displayed decimal or an execution failure. This aggregate comprises 111 records with different returned rows or values and two with runtime errors in every divergent scenario. Eleven additional records differ only in the last displayed decimal. Table~\ref{tab:output-impact-mechanisms} groups these observations by the mechanism exercised. The endpoint concerns agreement with the recomputed reference: a submission can remain unchanged while the reference answer changes.

Table~\ref{tab:output-impact-coverage} separates the remaining 69 records by their evidence status. The recorded checks for 57 concern publication-field granularity (8), version multiplicity (11), unobserved event types (10), observed timing (23), and value ranges (5). The other 12 involve task conditions, program fallbacks, or reference conventions. These checks did not establish a triggering event under the study criteria. Four of the 69 records do diverge in the two excluded scenarios. The findings therefore depend on the retained scenario criteria and do not establish invariance under all admissible changes. Among the 135 records with attempted execution, 113 (83.7\%) have a non-rounding discrepancy or execution failure, conditional on this screening.

\begin{table}[t]
\centering
\small
\setlength{\tabcolsep}{4pt}
\caption{Evidence status of the 193 annotated result-pass records. The last four rows comprise the 69 records without a retained divergence.}
\label{tab:output-impact-coverage}
\begin{tabular}{@{}p{0.83\columnwidth}r@{}}
\toprule
Evidence status & Records \\
\midrule
Non-rounding row or value discrepancy & 111 \\
Runtime failure only & 2 \\
Last-decimal difference only & 11 \\
Matched in retained scenarios & 8 \\
Only retained run timed out & 1 \\
Executed only in excluded scenarios & 2 \\
Data/code inspection without execution & 58 \\
\midrule
Total & 193 \\
\bottomrule
\end{tabular}
\end{table}

\begin{table*}[t]
\centering
\small
\setlength{\tabcolsep}{5pt}
\renewcommand{\arraystretch}{1.06}
\caption{Observed discrepancies under targeted data changes. Each record is assigned to one exercised mechanism. The columns separate row or value differences beyond the last displayed decimal, differences confined to that decimal, and records whose divergent scenarios all fail at runtime.}
\label{tab:output-impact-mechanisms}
\begin{tabular}{@{}p{0.60\textwidth}rrr@{}}
\toprule
Mechanism & \shortstack{Rows or\\values} & \shortstack{Last\\decimal} & \shortstack{Runtime\\only} \\
\midrule
Disclosure after the effective date & 28 & 0 & 1 \\
Recording or release after the cutoff & 39 & 11 & 0 \\
Listing or operating status around the research date & 22 & 0 & 1 \\
Later membership or classification applied to history & 7 & 0 & 0 \\
Missing latest value replaced by older data & 10 & 0 & 0 \\
Expired name selected & 1 & 0 & 0 \\
Later prices used to screen candidates & 4 & 0 & 0 \\
\midrule
Total & 111 & 11 & 2 \\
\bottomrule
\end{tabular}
\end{table*}

\paragraph{References and sensitivity.}
Three reference programs received temporal-selection corrections that preserve their snapshot answers before the affected scenarios were rerun. Seven records use study-specific reference variants, also matching the snapshot answers: three include the delisting-arrangement status in historical listing eligibility, and four preserve column names when the reference returns an empty table. The latter four scenarios place a missing financial value on entities for which that absence was not observed, although it occurs elsewhere in the table. Seven other records have discrepancy evidence only from excluding the delisting-arrangement status from listing eligibility, a convention on which the references differ. Excluding both sets of records leaves 102 of the original 193 (52.8\%) with retained discrepancy or execution-failure evidence.

One bond-selection record fails the unmodified-state comparison because of a basis-point column label, and that run's table was not retained. Its saved perturbed table retains a bond absent from the recomputed reference, so we retain it as a content discrepancy with the comparison exception recorded separately.

\paragraph{Label sensitivity.}
Two listing-eligibility conditions have disputed PIT status and were coded inconsistently: excluding the delisting-arrangement status after reconstructing historical status, and omitting a historical termination check. Treating both as non-PIT removes five records whose sole PIT reason is one of these conditions, leaving 188 of 380 result-pass records (49.5\%). Four additional records retain other PIT labels but have discrepancy evidence only from a disputed condition. Removing these witnesses leaves 104 of 188 records (55.3\%) with retained non-rounding discrepancy or execution-failure evidence. This is a mechanism-filtered count, not the total number of saved output mismatches. Conversely, treating both conditions as PIT and adding five identified unannotated passes gives 198 of 380 (52.1\%), with discrepancy or execution-failure evidence for 117 of 198 (59.1\%). Two identified result-fail cases raise that count from 345 to 347 of 520 (66.7\%). The search for omitted termination checks covered seven tasks, so these are results for the identified cases rather than exhaustive bounds. The primary results and agreement statistics retain the frozen labels.

\paragraph{Additional controls.}
We also execute 40 result-pass records without annotated A1--A3 errors on the tested tasks. Twenty-nine match the recomputed references in all tested scenarios, and four fail on an injected missing value because they do not handle nulls. Seven have content discrepancies: four involve the disputed listing conditions, one applies risk alerts without announcement dates, one includes a fund's expiry date in its valid interval, and one retains a release-month limit selected from the original snapshot. These mechanisms identify candidates for annotation review as well as sensitivity of a frozen program to changed data. The controls were selected from tasks with pass-side violations and do not estimate an audit miss rate. Scenario definitions, saved outputs, reference variants, and field-frequency checks are included in the evidence package.

\section{Supplementary Evaluation and Benchmark Comparisons}
\label{app:supplementary_comparisons}

\begin{figure*}[t]
\centering
\includegraphics[width=\textwidth,trim=0 40bp 0 65bp,clip]{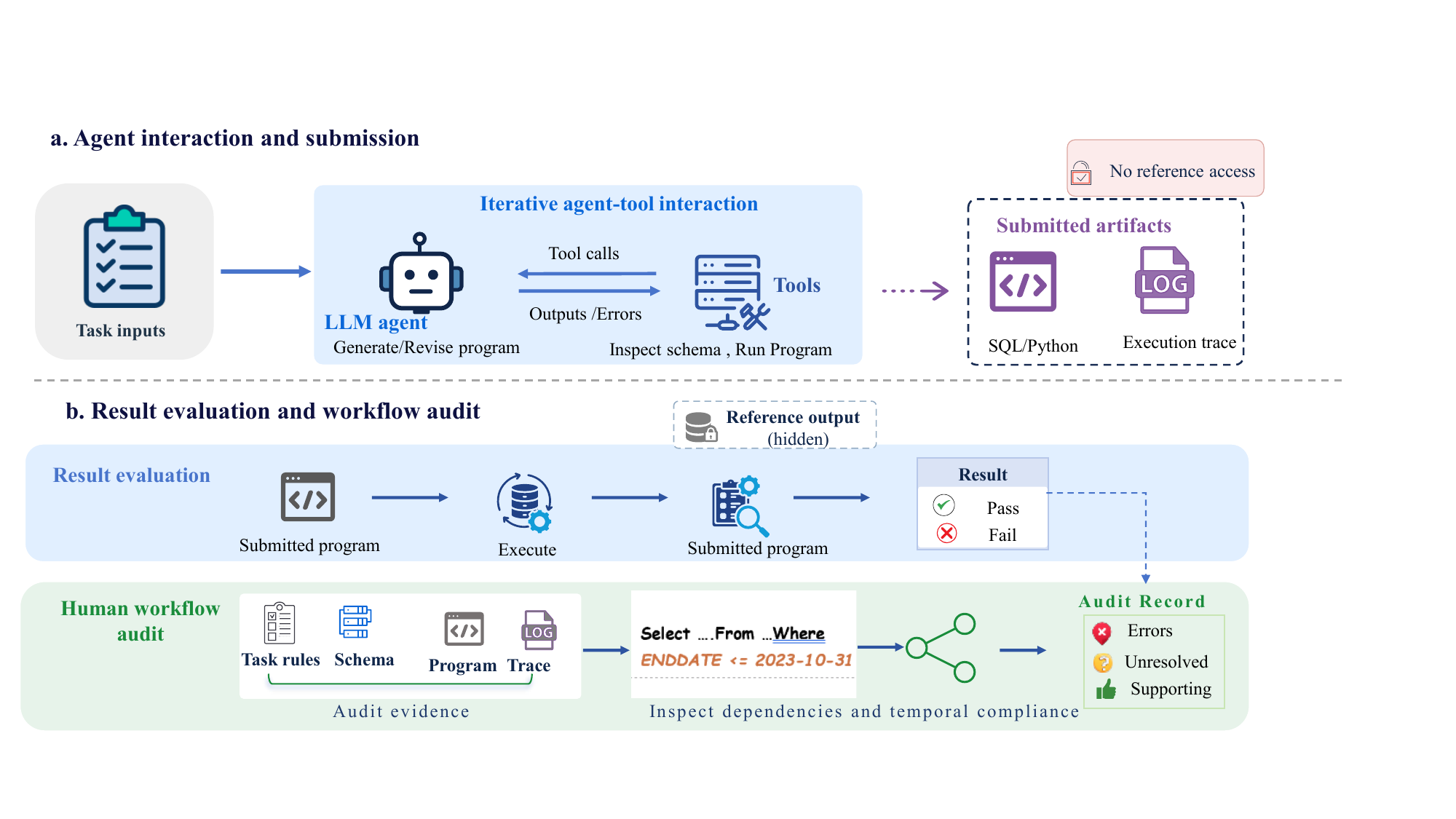}
\caption{Evaluation protocol of BuySideFlow. (a) Agent interaction and submission. (b) Result evaluation and human workflow audit.}
\label{fig:evaluation-protocol}
\end{figure*}

Figure~\ref{fig:evaluation-protocol} summarizes the evaluation protocol. Table~\ref{tab:benchmark-comparison} compares related benchmarks.

\subsection{PIT Guidance Comparison}
\label{sec:pit_guidance_comparison}

We compare two GPT-5.5/SWE-agent configurations: \emph{PIT-augmented}, with PIT implementation guidance in the prompts and business-rule files, and \emph{No PIT guidance}, with this guidance removed. PIT-augmented achieves 35.05\% accuracy (143/408), compared with 26.23\% (107/408) for No PIT guidance, a difference of 8.82 percentage points.

\begin{figure}[!htbp]
\centering
\includegraphics[width=\columnwidth]{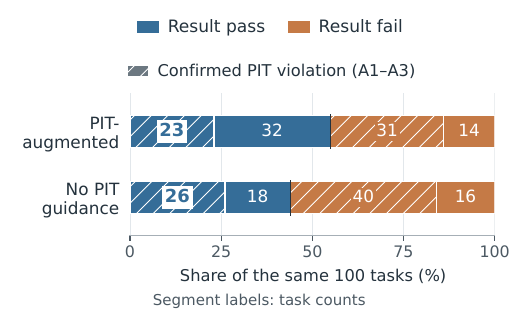}
\caption{Joint distribution of result outcomes and confirmed PIT violations on the same 100 audited tasks. Colors encode result outcomes; hatching marks PIT violations (A1--A3). Segment labels are task counts, with result-pass/PIT-violation counts highlighted. Unhatched segments indicate no confirmed violation, which does not establish PIT compliance.}
\label{fig:pit-guidance-audit}
\end{figure}

We also examine a paired expert audit of 100 shared tasks under the same task-specific temporal requirements. Figure~\ref{fig:pit-guidance-audit} shows the joint distribution of result outcomes and confirmed PIT violations. On this set, result accuracy is 55\% for PIT-augmented and 44\% for No PIT guidance, while confirmed PIT violations occur in 54\% and 66\% of submissions, respectively.

Within the 100 task pairs, removing PIT guidance coincides with 16 results changing from pass to fail and five from fail to pass. No PIT guidance has confirmed PIT violations on 17 tasks with no confirmed violation under PIT-augmented, while the reverse holds for five, giving a net increase of 12 percentage points. A1 (43 vs. 57) and A2 (6 vs. 19) are more frequent under No PIT guidance, whereas A3 (27 vs. 19) is less frequent; each comparison lists PIT-augmented first. Other prompt differences limit attribution of these changes to PIT guidance.

Among result-pass submissions, expert audit confirms PIT violations in 23/55 (41.8\%) for PIT-augmented and 26/44 (59.1\%) for No PIT guidance. These proportions describe each configuration's own passing submissions. Result matching therefore leaves temporal violations undetected under both configurations, including when explicit PIT guidance is provided.

\begin{table*}[!ht]
\centering
\newcommand{\rwcheck}{{\usefont{U}{msa}{m}{n}\char88}}
\caption{Related benchmarks for database analysis, financial tool use, and quantitative investment. \protect\rwcheck{}: specified temporal control or check; NR: not reported in the cited work.}
\label{tab:benchmark-comparison}
\small
\setlength{\tabcolsep}{3pt}
\renewcommand{\arraystretch}{1.08}
\begin{tabular}{@{}>{\raggedright\arraybackslash}p{0.14\linewidth}>{\raggedright\arraybackslash}p{0.15\linewidth}>{\raggedright\arraybackslash}p{0.23\linewidth}>{\raggedright\arraybackslash}p{0.205\linewidth}>{\raggedright\arraybackslash}p{\dimexpr0.275\linewidth-8\tabcolsep\relax}@{}}
\toprule
Benchmark & Task setting & Data and execution & Output evaluation & Temporal checks \\
\midrule
BULL & Financial querying & Financial databases; generated SQL & Execution-result accuracy & NR \\
\addlinespace[3pt]
Spider~2.0 & Enterprise analytics & Databases and code projects; SQL/Python/dbt & Task-specific output checks & NR \\
\addlinespace[3pt]
DAB & Data analysis & Databases; SQL and Python tools & Deterministic answer checks & NR \\
\midrule
FinGAIA & Financial analysis & Web, files, and Python & Answer accuracy & NR \\
\addlinespace[3pt]
FinSkillBench & Investment subtasks & Market/XBRL retrieval; condition-dependent financial tools & Structured outputs; deterministic checks & \rwcheck{} As-of data slices \\
\midrule
AlphaQT-Bench & Factor generation & Price/volume data; single-turn Python generation & Reference-output accuracy & \rwcheck{} Prefix-causality safety gates functional accuracy \\
\addlinespace[3pt]
OpenPM & Portfolio allocation & Market and vintage data & Portfolio performance and rule adherence & \rwcheck{} Runtime availability audit and certificate \\
\midrule
\textbf{BuySideFlow} & Chinese buy-side research & Fixed financial database; agent-written SQL or SQL--Python & Reference checks for tables, figures, and mixed outputs & \rwcheck{} Separate expert audit of executed dependencies: cutoffs, historical states, and record versions \\
\bottomrule
\end{tabular}
\end{table*}

\subsection{Additional Related Work}
Chinese text-to-SQL benchmarks assess language and schema grounding \citep{min2019pilot,wang2020dusql}. Distilled test suites strengthen SQL evaluation across database instances \citep{zhong2020semantic}.

Financial benchmarks cover document-based numerical reasoning \citep{chen2021finqa,zhu2021tat}, filing research \citep{bigeard2025finance}, executable trading strategies \citep{khoroshilov2026quantcode}, and factor generation \citep{luo2026alphabench,luo2026alphaqt}. QuantEval's trading-code subset compares backtest metrics and prohibits future-information access \citep{kang2026quanteval}.

Temporal databases distinguish valid from transaction time \citep{jensen1999temporal}. Real-time datasets preserve releases and revisions \citep{croushore2001realtime}. Formal look-ahead analysis establishes temporal non-interference for a restricted language under trusted availability declarations \citep{fonseca2026lookahead}.

\end{document}